\documentclass{article} 
\usepackage{iclr2027_conference,times}

\usepackage{amsmath,amsfonts,bm}

\def\eqref#1{equation~\ref{#1}}

\def\1{\bm{1}}

\DeclareMathAlphabet{\mathsfit}{\encodingdefault}{\sfdefault}{m}{sl}
\SetMathAlphabet{\mathsfit}{bold}{\encodingdefault}{\sfdefault}{bx}{n}

\usepackage{hyperref}
\usepackage{url}
\usepackage{graphicx}
\usepackage{booktabs}
\usepackage{enumitem}
\usepackage{float}

\title{ACTOR: Action-Centered Temporal Organization for Literary Role-Playing Agents}

\author{
Yuxiang Gao$^1$ \quad Tianjie Ju$^2$\thanks{Corresponding author.} \quad Zheng Wu$^3$ \quad Zongru Wu$^{3,2}$ \quad Hao Fei$^4$\\
$^1$Harbin Institute of Technology \\
$^2$National University of Singapore \\
$^3$Shanghai Jiao Tong University \\
$^4$University of Oxford 
}

\iclrfinalcopy
\begin{document}

\maketitle

\begin{abstract}
Large language agents for literary role-playing are typically modeled around dialogue generation, leaving characters' non-dialogue behaviors and their dynamic evolution over story time underrepresented. In this paper, we propose  \textbf{ACTOR} (\textbf{A}ction-\textbf{C}entered \textbf{T}emporal \textbf{O}rganization for Literary \textbf{R}ole-Playing Agents), an automated framework for constructing such behavior trajectories from raw novel texts. ACTOR extracts dialogue and persona-relevant non-dialogue behaviors, represents each behavior using a Thought–Action–Observation structure, and organizes behavior trajectories according to the internal temporal order of the story. We build a behavior dataset from 143 diverse literary works. For each source behavior, we generate counterfactual behaviors that are contextually plausible yet inconsistent with the character’s persona. 
We find that existing models struggle to reliably infer character-consistent actions from historical behaviors and evolving character states.
Using these behavior pairs, we train models with Regularized Preference
Optimization (RPO), which augments DPO with a negative log-likelihood loss on the source behavior to jointly optimize behavior preference and generation. We further design a four-way next-behavior selection task to evaluate the model’s ability to faithfully mimic character actions. Across 17,560 evaluation instances, the model trained on ACTOR’s preference pairs achieves substantially improved accuracy. This result demonstrates that structured, temporally organized supervision enables models to leverage character profiles and prior behaviors to infer subsequent actions, greatly advancing literary role-playing beyond mere dialogue imitation. Code and data are available at https://github.com/SilentStar233/ACTOR.
\end{abstract}

\section{Introduction}
\label{sec:intro}

As research on role‑playing language agents advances, researchers have constructed richer role representations and high-quality training datasets~\citep{wang2026roleplayingagentsdrivenlarge}. Early studies draw on diverse textual sources such as scripts, biographies, and novels to build faithful role representations. These sources support retrieval, fine-tuning, and generation constraints, yielding text-grounded character profiles and literary worlds~\citep{shao-etal-2023-character,ran-etal-2025-bookworld}. Recent works further dynamically update character profiles and memories to simulate evolving character states across different story stages, effectively improving the faithfulness of role‑playing~\citep{xiao2026dream,peng-etal-2026-deriving}.

However, fine-grained timestamp-aware behavior modeling cannot be adequately supported by these representations. Some prior works treat actions as embedded components of dialogue messages and thus mainly serve dialogue generation~\citep{wang2025coser}. Other existing lines of research predict characters’ next actions based on plot summaries or pre‑constructed coarse‑grained action sequences~\citep{xu-etal-2025-character,song2026arcaneroleplayinglanguageagents}. Consequently, literary‑oriented RPLA still lacks character‑centric atomic‑level behavior trajectories. Meanwhile, existing methods generally organize character experiences following the narrative order of source texts, with little explicit distinction between the actual chronological order of events and their disclosure order in original literary works.

To address the above practical challenges, we propose a pipeline framework that automatically constructs character-behavior trajectories from full-length novels. Behaviors extracted by this framework include both verbal dialogues and non-dialogue actions, and the trajectories are organized strictly according to the actual chronological order of unfolding events, providing a unified data foundation for investigating character behaviors and their gradual evolution throughout the story progression. During dataset construction, we concurrently generate contextually plausible yet persona-mismatched counterfactual actions for each ground-truth behavior to support subsequent model training and evaluation. Finally, we build our curated dataset extracted from 143 literary works.

For model training, we select 115 works from the curated corpus and use the first \(80\%\) of each character’s behavior trajectory for training. As each source behavior is paired with a contextually plausible but persona‑inconsistent counterfactual behavior, we use Regularized Preference Optimization (RPO), a joint preference-and-generation objective combining Direct Preference Optimization (DPO) with a negative log-likelihood loss on the source behavior. We treat the source behavior as chosen and one counterfactual behavior as rejected. These signals train the model to infer and generate actions consistent with the character's current state and behavioral prior history in context.

\begin{figure}[t!]
\begin{center}
\includegraphics[width=1.0\textwidth]{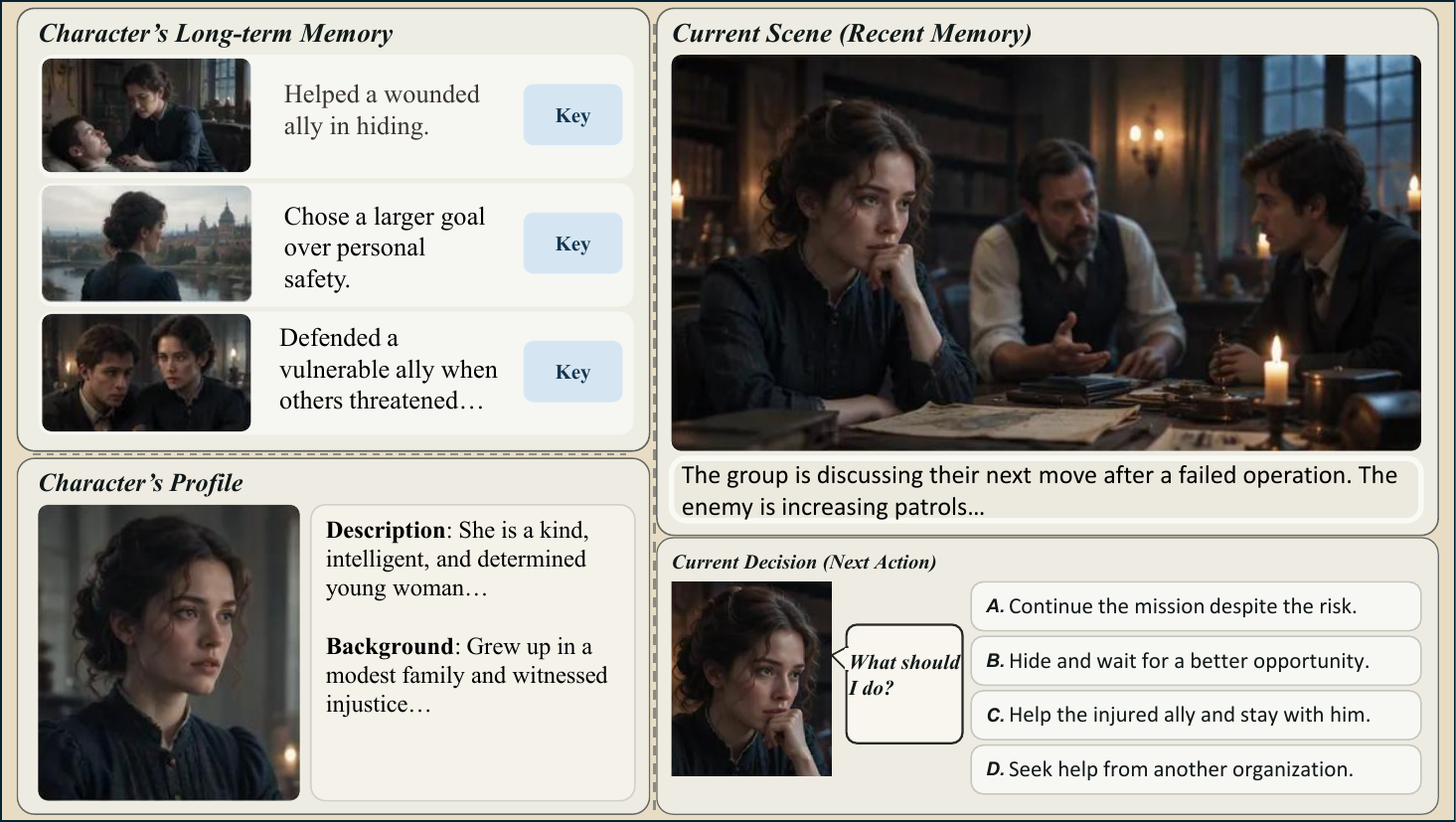} 
\end{center}
\caption{Overview of our behavior-modeling evaluation. Given character-related information, the model needs to select the most plausible next action from four options.} 
\label{fig:evaluation} 
\end{figure}

To evaluate the model’s behavior‑modeling capacity, we take the subsequent \(20\%\) of behaviors from each character trajectory within the aforementioned 115 works as temporal evaluation data. The remaining 28 unseen works excluded from training are entirely reserved for cross‑book evaluation. For each evaluation instance, given the character profile, historical memory and current scene, the model is required to select the most plausible next action taken by the character among one ground‑truth behavior and three distinct counterfactual behaviors. The former setup examines whether the model can continue observed character experiences, whereas the latter tests the model’s ability to predict behaviors based on structured character information without access to the source literary works.

The main contributions of this paper are summarized as follows:
\begin{itemize}[noitemsep, topsep=1pt,leftmargin=*]
    \item We propose ACTOR, an automated framework that extracts character-centered
    atomic behaviors from raw literary texts and organizes behavior trajectories
    according to their actual occurrence time in the story. ACTOR addresses the
    lack of character behavior data in existing resources. 

    \item We formulate literary role-playing as an action-centered temporal modeling
    task: at a specified point in the story, the model predicts a character's
    next behavior based on the character's state and previous behavior history.
    This task extends literary role-playing beyond dialogue response generation to
    the unified modeling of dialogue and non-dialogue behaviors.

    \item We construct a next-behavior selection benchmark to objectively evaluate a
    model's ability to imitate behavior based on character information, and use
    the behavior data generated by ACTOR to train models. Experiments on this
    benchmark show that training with these data improves a model's ability to
    imitate the behavior of a specified character.
\end{itemize}

\section{Related Work}
\label{sec:related_work}

\subsection{Literary Character Role‑Playing Grounded in Textual Sources}

Early role‑playing systems primarily ground language models in character descriptions or historical dialogue
(\citealp{chen2024from}; \citealp{chen2025oscarsaitheatersurvey}).~\citet{li2023chatharuhirevivinganimecharacter} retrieve character-specific dialogue from scripts to better support response generation.~\citet{shao-etal-2023-character} reconstruct character experiences from biographical descriptions and train a dedicated agent for each character.~\citet{wang-etal-2024-rolellm} extend role-conditioned instruction tuning to a broader range of diverse roles.~\citet{lu-etal-2024-large} generate and refine role-playing supervision through self-alignment.

Later studies construct richer character information from literary sources~\citep{park-etal-2025-charactergpt,electronics14193859}.~\citet{wang2025coser} extract novel‑sourced authentic conversations alongside scene context and character experiences, augmenting dialogue turns with associated actions and thoughts.~\citet{cheng-etal-2026-psymem} leverage fine‑grained psychological attributes to organize character memory for improved long‑term consistency.~\citet{du-etal-2026-human} reverse‑synthesize character reasoning from literary dialogues and apply reinforcement learning to enhance character motivation modeling.~\citet{tang2026stayingcharacterperspectiveboundedmemory} structure book‑derived memory from the character’s egocentric perspective to prevent god‑view information leakage. These works enrich role-playing information, yet they are optimized for dialogue response generation rather than sequential behavior prediction.

\subsection{Temporal Character Modeling via Action Trajectories}

Another line of research examines characters at specific points in a narrative~\citep{ahn-etal-2024-timechara,xu-etal-2025-character}. \citet{ahn-etal-2024-timechara} require models to respect chapter-specific knowledge boundaries and avoid referring to future events that remain unrevealed within the current narrative segment. \citet{xu-etal-2025-character} reconstruct key decisions from a character's prior experiences. \citet{yuan-etal-2024-evaluating} summarize character development from complete fictional works and evaluate whether the resulting profiles support character understanding. These studies show that character behavior depends not only on general personality, but also on experiences accumulated before the current story point.

Recent studies model character evolution based on plot progression and personal experiences~\citep{yang2025evolvtripenhancingliterarycharacter,peng2025codifying}.~\citet{xiao2026dream} build temporally‑ordered causal event graphs over novels for point‑specific character profiles, while~\citet{zong2026evolvingworldopenschemaframeworkcoevolving} co‑update character states and fictional worlds from dialogue and narrative text.~\citet{song2026arcaneroleplayinglanguageagents} assess agent behavior across psychological phases of character development.~\citet{peng-etal-2026-deriving} learn context‑dependent behavioral rules from Fandom‑sourced character‑labeled action sequences and scene–action pairs, and~\citet{peng2026bookmarksefficientactivestoryline} encode storylines as continuous actions with task‑relevant history for next‑action prediction.
However, neither line of work extracts complete character action trajectories from novels nor reconstructs the genuine chronology of character experiences.

\section{Introduction of ACTOR}
\label{sec:data and train}

\subsection{Dataset Construction}
\label{sec:dataset_construct}

\begin{figure}[t!]
\begin{center}
\includegraphics[width=1.0\textwidth]{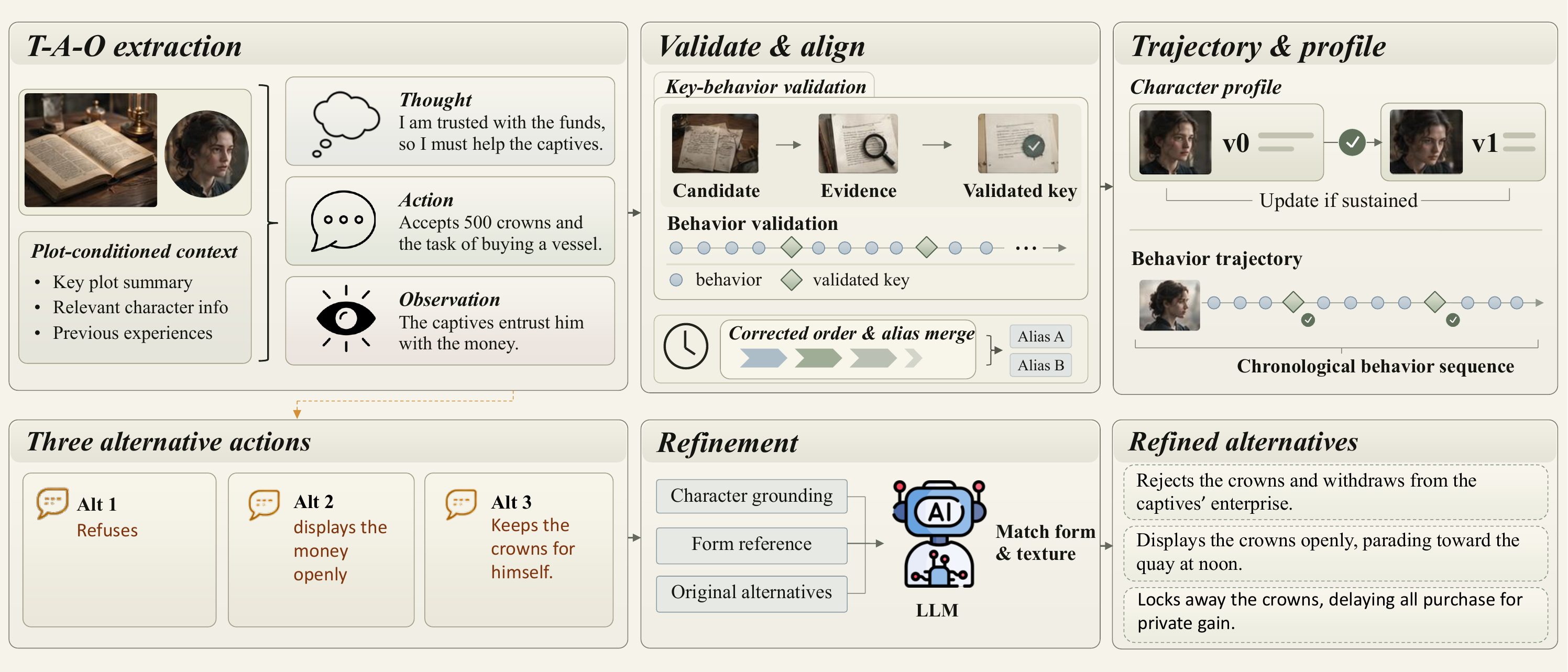} 
\end{center}
\caption{ACTOR dataset construction pipeline. Plot-conditioned context is transformed into Thought–Action–Observation (T-A-O) units, followed by key-behavior validation, temporal and entity alignment, and dynamic profile updates for sustained changes. Each behavior is paired with three situation-plausible but target-character-inconsistent alternatives, which are refined to match the real action’s form and texture while leaving the real action unchanged.} 
\label{fig:pipeline} 
\end{figure}

\paragraph{T-A-O behavior extraction}
We define character behavior as contextual utterances and distinctive non-dialogue actions in literary scenarios. Deliberate silence and avoidance are treated as valid behaviors, whereas routine actions that reflect no unique personal traits or active choices are filtered out. Each behavior follows a Thought--Action--Observation (T-A-O) structure: Thought describes the internal mental state and decision rationale before an action, Action corresponds to the verbal utterance or physical conduct, and Observation captures subsequent interpersonal reactions and environmental changes perceptible to the character. To efficiently manage long-text contexts, the extraction pipeline adopts a hierarchical granularity fallback, processing full plots first, followed by individual scenes and character-level contexts as a last resort.

\paragraph{Counterfactual alternatives}
For each behavior, the system generates alternative actions that are plausible in the current situation but inconsistent with the target character. So selecting the correct behavior requires genuine understanding of the character. These alternatives are generated together with the extracted behaviors and are refined after constructing character trajectories and profiles.

\paragraph{Key-behavior validation}
As shown in the validation module of Figure~\ref{fig:pipeline}, we clearly distinguish ordinary and key behaviors. A key behavior must satisfy two core criteria: the character has reasonable alternative options at the moment, and the selected action alters persistent states or future situational possibilities. To reduce annotation noise from single-pass generation, the system first produces candidate key behaviors, identifies scenes in subsequent plots that reflect long-term behavioral impacts, and verifies the candidates through an independent evidence-checking procedure. Only evidence-supported candidates are retained as valid key behaviors.

\paragraph{Temporal alignment}
After behavior verification, we re-annotate the precise occurrence time of each scene using the original plot text. The textual reading order is used as the default timeline, while ordering anomalies potentially caused by flashbacks or reversed narration are detected and converted into preliminary adjustment proposals. Each proposal is independently checked against the source text, and only scene reorderings supported by sufficient textual evidence are applied. Character aliases and scattered mentions are then normalized to construct complete character-level behavioral trajectories in the corrected in-story chronological order.

\paragraph{Character data}
The resulting behavior trajectory of a character $c$ is represented as:
\begin{equation}
\mathcal{T}_c
=
\left(
b_1,\ldots,b_{N_c}
\right),
\qquad
b_i
=
\left(
s_i,h_i,a_i,o_i,r_i,t_i,\kappa_i
\right),
\quad
t_1 \leq \cdots \leq t_{N_c},
\label{eq:behavior_trajectory}
\end{equation}
where $s_i$ denotes the scene context, $(h_i,a_i,o_i)$ is the Thought--Action--Observation representation, $r_i$ is the textual position, $t_i$ is the reconstructed story-time position, and $\kappa_i\in\{0,1\}$ indicates whether the behavior is a valid key behavior. The trajectory is ordered by story time, while the textual position is retained for source tracing.

Based on each trajectory, initial character profiles are derived from descriptions before the character's first appearance. Each valid key behavior triggers a profile-update check over all experiences accumulated after the latest profile version. A new profile version is saved only when these experiences lead to sustained character changes.

\paragraph{Alternative refinement}
Finally, the initially generated counterfactuals are refined to match the length and structural form of real behaviors while remaining situationally reasonable and inconsistent with the target character. We then anonymize all datasets. Book titles and character names are withheld from the model during both training and evaluation to prevent leakage from the base model's prior knowledge and improve generalization in character-action modeling.

\subsection{Action Preference Training}
\label{action_training}

Based on the character behavioral trajectories, we formulate character action prediction as a conditional preference learning task. The input context $\textbf{x}$ consists of four types of inputs before each target action: a dynamic profile of the current character state, long-term memory of validated key behaviors, short-term memory of recent scene interactions, and situational context containing background, temporal cues, and participant information. All inputs are strictly sampled from pre-action story content, ensuring the model only uses information accessible to the character at that moment.

In terms of training paradigm, our dataset is naturally suited to preference optimization. For the same prediction context, we possess both the character's genuine behavior $\mathbf{y^{+}}$ extracted from the source text and counterfactual behaviors with matched length and format but inconsistent character profiles, which form an explicit relative preference relation: the genuine behavior should be preferred over actions the character would not take. Among the three counterfactual behaviors, the one closest in length to the genuine behavior is selected as $\mathbf{y^{-}}$ to minimize confounding effects from output-length differences. The source behavior and selected counterfactual are stored as chosen and rejected, respectively, defining the preference relation $\mathbf{y^{+}} \succ \mathbf{y^{-}}$.

Direct Preference Optimization (DPO) learns this relation by optimizing the relative likelihood of the two behaviors against a fixed reference policy~\citep{rafailov2023dpo}:
\begin{equation}
\mathcal{L}_{\mathrm{DPO}}(\theta)
=
-\mathbb{E}_{(x,y^{+},y^{-})\sim\mathcal{D}}
\left[
\log \sigma
\left(
\beta
\left(
\log
\frac{\pi_{\theta}(y^{+}\mid x)}
     {\pi_{\mathrm{ref}}(y^{+}\mid x)}
-
\log
\frac{\pi_{\theta}(y^{-}\mid x)}
     {\pi_{\mathrm{ref}}(y^{-}\mid x)}
\right)
\right)
\right],
\label{eq:dpo}
\end{equation}
where $\pi_{\theta}$ is the policy being optimized, $\pi_{\mathrm{ref}}$ is the fixed reference policy, and $\beta$ controls the scaling of the preference objective. 

However, this relative objective does not explicitly require the likelihood of the source behavior itself to increase. Consequently, it may only widen the preference margin during action selection, without increasing the probability of directly generating utterances or actions consistent with the target character.  Following \citet{liu2024rpo}, we therefore adopt RPO, which augments the DPO objective with a token-level negative log-likelihood loss on the source behavior:
\begin{equation}
\mathcal{L}_{\mathrm{NLL}}(\theta)
=
-\mathbb{E}_{(x,y^{+})\sim\mathcal{D}}
\left[
\frac{1}{|y^{+}|}
\sum_{t=1}^{|y^{+}|}
\log \pi_{\theta}
\left(
y^{+}_{t}\mid x,y^{+}_{<t}
\right)
\right].
\label{eq:chosen_nll}
\end{equation}

The final training objective is:
\begin{equation}
\mathcal{L}_{\mathrm{RPO}}(\theta)
=
\mathcal{L}_{\mathrm{DPO}}(\theta)
+
\alpha\mathcal{L}_{\mathrm{NLL}}(\theta),
\label{eq:rpo}
\end{equation}
where $\alpha$ controls the contribution of direct supervision on the source behavior. We set $\beta=0.1$ for all models and use $\alpha=1.0$ for the Qwen-2.5 models and $\alpha=0.25$ for Llama-3.1-8B. The resulting objective trains the model both to distinguish character-consistent behaviors from counterfactuals and to increase the likelihood of
the source behavior itself.

\section{Evaluation}
\label{evaluation}

\subsection{Evaluation Setup}
\label{sec:evaluation_setup}

\paragraph{Evaluated Models}
We evaluate six pretrained models as baselines: Qwen-2.5-Instruct 7B and 14B~\citep{qwen2025qwen25technicalreport},
Llama-3.1-Instruct 8B~\citep{grattafiori2024llama3herdmodels},
Qwen-3-32B~\citep{yang2025qwen3}, CoSER-8B~\citep{wang2025coser}, and HER-32B model~\citep{du-etal-2026-human}. These baselines cover general-purpose instruction-tuned large language models and existing specialized role-playing models for literary character simulation. We further evaluate three ACTOR models trained with RPO: ACTOR-Qwen-2.5-7B, ACTOR-Qwen-2.5-14B, and ACTOR-Llama-3.1-8B.

\paragraph{Evaluation Data}
The evaluation data are directly extracted from character behavior trajectories constructed from 143 literary works. For 115 works that contribute to
training, the first $80\%$ of each character trajectory is used to construct the training data, while the remaining $20\%$ is reserved for temporal evaluation. The other 28 works are excluded from training and are used entirely for cross-work evaluation. The final evaluation set contains
17,560 test instances, including 11,770 temporal-evaluation instances and 5,790 cross-work instances.

\paragraph{Task and Metrics}
We formulate evaluation as a four-way next-behavior selection task. Given
the character profile, behavioral memories, and current scene before the
target behavior, the model must select the candidate behavior most likely
to be performed by the target character. Each instance contains one
source-text behavior and three counterfactual candidates. Selection accuracy is the primary metric. We additionally report accuracy on temporal and cross-work subsets, as well as on key and non-key behavior subsets, to examine model performance across different settings.

\paragraph{Open-ended Generation Evaluation}
Beyond next-behavior selection, we adopt the CoSER role-playing evaluation
framework~\citep{wang2025coser} to examine open-ended character interaction. CoSER evaluates models through multi-turn scenarios and combines environment simulation, next-state prediction, and language-model-based judging. It provides scores along four core dimensions: plot consistency, human-likeness, character fidelity, and plot quality. We use this evaluation to test whether ACTOR-derived training transfers from selecting character-consistent behaviors to generating coherent actions in dynamic interactive story settings.

\subsection{Main Results}

\begin{table*}[t]
\caption{Main results on the complete 17,560-instance evaluation set under
the anonymous setting. All values are selection accuracy
(\%). Bold and underlined values denote the best and second-best results,
respectively.}
\label{tab:main_results}
\centering
\resizebox{0.8\textwidth}{!}{%
\begin{tabular}{lccccc}
\toprule
Model & Overall & Temporal & Cross-work & Key & Non-key \\
\midrule
Qwen-2.5-Instruct 7B
& 71.34 & 71.29 & 71.43 & 71.10 & 71.55 \\

Qwen-2.5-Instruct 14B
& 76.96 & 77.04 & 76.79 & 76.85 & 77.05 \\

Llama-3.1-Instruct 8B
& 69.15 & 68.45 & 70.59 & 69.97 & 68.43 \\

Qwen-3-32B
& 78.30 & 77.97 & 78.96 & 79.14 & 77.56 \\

CoSER-8B
& 57.41 & 56.89 & 58.46 & 60.56 & 54.63 \\

HER-32B
& 65.65 & 65.02 & 66.93 & 67.24 & 64.25 \\

\midrule
ACTOR-Qwen-2.5-7B
& \underline{80.20} & \underline{80.23} & \underline{80.14}
& 80.21 & \underline{80.19} \\

ACTOR-Qwen-2.5-14B
& \textbf{86.00} & \textbf{86.16} & \textbf{85.68}
& \textbf{86.40} & \textbf{85.65} \\

ACTOR-Llama-3.1-8B
& 78.04 & 78.24 & 77.63 & \underline{79.37} & 76.87 \\

\bottomrule
\end{tabular}%
}
\end{table*}

Table~\ref{tab:main_results} reports selection accuracy on the evaluation set under the anonymous setting.

All three ACTOR models outperform their corresponding backbones, with overall gains ranging from 8.86 to 9.04 percentage points. ACTOR-Qwen-2.5-14B achieves the highest overall accuracy of 86.00\%, while ACTOR-Qwen-2.5-7B reaches 80.20\% and surpasses the larger Qwen3-32B baseline.
\textbf{These results clearly show that ACTOR improves character-consistent behavior selection across diverse model families and sizes.} Since all models are evaluated under the same strictly anonymous setting, the gains cannot be attributed to access to book titles or original character names. The comparison with Qwen3-32B further suggests that model scale alone does not substitute for targeted behavioral supervision tailored for literary character modeling.

The cross-work subset contains literary works excluded from training,
making it a test of whether the learned behavior patterns transfer beyond
familiar stories. \textbf{ACTOR retains substantial gains on these unseen
works:} all three models improve over their backbones by 7.04--8.89
percentage points. The two Qwen-based models show similar gains on the
temporal and cross-work subsets, whereas the Llama-based model has a smaller
cross-work gain. These improvements are unlikely to arise solely from
continuing trajectories seen during training. Instead, the learned
behavioral supervision generalizes effectively to unseen characters and
narrative scenarios.

The improvements extend to both key and non-key behaviors. For example,
ACTOR-Qwen-2.5-7B gains 9.11 percentage points on key behaviors and 8.64
points on non-key behaviors; the other two ACTOR models show similarly
balanced gains. \textbf{ACTOR therefore improves the selection of ordinary
persona-relevant behaviors as well as salient plot events.} This matters
for role-playing, where character consistency must also be maintained in
less consequential, everyday interactions.

CoSER-8B and HER-32B obtain lower scores on this selection benchmark
despite being developed for role-playing. \textbf{Their results suggest
that training for open-ended role-playing does not necessarily yield
strong fine-grained action selection.} The difference may reflect the
objectives of these models and should not be taken as evidence of weaker
role-playing ability overall. We therefore examine open-ended interaction
separately using the CoSER framework.

\subsection{Qualitative Case Study}
\label{sec:case_study}

To complement the quantitative results, we present three illustrative instances in which ACTOR-Qwen-2.5-7B selects the behavior consistent with the source text, whereas its backbone selects a counterfactual behavior. As shown in Table~\ref{tab:case_study}, these cases require different forms of behavioral reasoning: tracking a character's long-term personality transformation, maintaining local dialogue coherence, and adapting an ongoing action to changes in the current scene. For readability, the contexts shown in the main text are manually summarized; more detailed versions of the three instances are provided in Tables~\ref{tab:case_detail_state}--\ref{tab:case_detail_action}
in Appendix~\ref{app:case_details}.

\begin{table*}[t]
\caption{Representative cases from the evaluation results. Only
decision-relevant evidence is shown.}
\label{tab:case_study}
\centering
\resizebox{\textwidth}{!}{%
\begin{tabular}{p{0.12\textwidth}p{0.31\textwidth}p{0.22\textwidth}p{0.22\textwidth}}
\toprule
Case
& Relevant pre-action evidence
& Qwen-2.5-7B-Instruct
& ACTOR-Qwen-2.5-7B / Source behavior \\
\midrule

State transformation
&
Formerly cold and dismissive of poverty, the character later resolves to
change and repeatedly behaves with kindness and generosity. He only appears
severe when the clerk arrives late.
&
Punishes the clerk and refuses further support for his family.
&
Raises the clerk's salary, supports his family, and asks him to add coal to
the fire. \\

\midrule
Dialogue continuity
&
The profile lists only the character's occupation. In the conversation, she
alternates between teasing disclosures and assurances about another person's
visits.
&
Promises to disclose the information: ``I'll tell.''
&
Reassures the interlocutor that she will not disclose it: ``Certainly not.'' \\

\midrule
Action adaptation
&
After declaring his willingness to kill an opponent and drawing his sword,
the character is challenged by another character who intervenes.
&
Abandons the attack and shields the original target.
&
Redirects the attack toward the intervening character. \\

\bottomrule
\end{tabular}%
}
\end{table*}

The first case concerns a character whose behavior has changed over time. The
backbone selects a punitive action, which is superficially plausible given the
character's earlier reputation for coldness. However, the behavioral history
shows he has recognized the consequences of his former conduct and repeatedly acted with kindness and generosity. His severe manner at the
beginning of the current interaction is therefore performative rather than
evidence of a return to his former state. ACTOR instead selects the
source-consistent action of raising the clerk's salary and supporting his
family. Thus, ACTOR can \textbf{prioritize a temporally updated character state over an earlier salient trait}.

In the second case, the character profile provides little useful personality
information beyond the character's occupation. The decision must therefore be
inferred primarily from the ongoing conversation, which alternates between
direct answers, teasing disclosures, and assurances. The backbone predicts an
eager promise to disclose the information, whereas ACTOR selects the brief
assurance appearing in the source text. This example suggests that ACTOR
can \textbf{use the sequential organization of recent behaviors when profile cues are
insufficient to determine the next action}.

The third case highlights ACTOR's ability to maintain action continuity
after a change in the scene. The character has already drawn his sword
against an opponent when another character intervenes. ACTOR redirects the
attack toward the intervening character, preserving the character's existing
intent while responding to the new situation. In contrast, the backbone
incorrectly protects the original target, which conflicts with both the preceding attack
and the character's stated objective. The contrast indicates that effective
prediction requires \textbf{preserving the character's initial intent while updating
the immediate action in response to the intervention}.

\subsection{Open-ended Generation Results}
\label{sec:open_ended_results}

\begin{table*}[t]
\centering
\caption{Open-ended generation results under the CoSER evaluation
framework.}
\label{tab:coser_results}
\resizebox{\textwidth}{!}{%
\begin{tabular}{lccccc}
\toprule
Model
& Plot Consistency
& Human-likeness
& Character Fidelity
& Plot Quality
& Average \\
\midrule
CoSER-8B
& 16.07 & 6.96 & 30.83 & 25.31 & 19.79 \\
\midrule
Qwen-2.5-Instruct 7B
& 11.70 & 1.17 & 21.87 & 6.97 & 10.43 \\
ACTOR-Qwen-2.5-7B
& \textbf{22.57} & \textbf{3.88} & \textbf{41.12} & \textbf{20.40} & \textbf{21.99} \\
\midrule
Qwen-2.5-Instruct 14B
& 25.30 & 3.90 & \textbf{42.95} & 26.60 & 24.69 \\
ACTOR-Qwen-2.5-14B
& \textbf{25.30} & \textbf{15.00} & 38.50 & \textbf{34.75} & \textbf{28.39} \\
\midrule
Llama-3.1-Instruct 8B
& 6.35 & 2.80 & 16.40 & 3.85 & 7.35 \\
ACTOR-Llama-3.1-8B
& \textbf{16.75} & \textbf{7.65} & \textbf{21.10} & \textbf{23.80} & \textbf{17.33} \\
\bottomrule
\end{tabular}%
}
\end{table*}

To verify whether ACTOR can generalize beyond candidate behavior selection, we test the models on open-ended multi-turn character interaction using the CoSER evaluation framework. GPT-5.6-Luna serves as both helper and judge model across all experiments. Since our experimental setup differs from the original CoSER work, we compare models under a unified evaluation protocol instead of directly matching our absolute scores against those reported in the original paper.

CoSER is constructed from multi-character dialogues extracted from authentic literary works, where each message integrates speech, physical actions, and inner thoughts. It formulates character-specific supervised training by taking scene context and other characters’ utterances as input to predict the target character’s responses. Additionally, it optimizes environment modeling and next-speaker prediction to facilitate multi-agent simulation. Such data structures and training objectives fundamentally differ from ours, yet our models still achieve performance gains.

As shown in Table~\ref{tab:coser_results}, all three ACTOR models achieve higher overall scores than their corresponding backbone models, with particularly clear improvements for ACTOR-Qwen-2.5-7B and ACTOR-Llama-3.1-8B. The improvements are not uniform across individual dimensions because ACTOR is not trained on CoSER-specific interaction procedures or output formats. Nevertheless, the consistent improvement in overall performance suggests that, even without explicit training on CoSER data, \textbf{preference learning over temporally structured character trajectories can enable models to exhibit features associated with character dialogue style in open-ended interaction.} These results also demonstrate ACTOR’s broader application potential. Details are shown in appendix~\ref{app:coser_examples}.

\subsection{Ablation Study}
\label{sec:ablation}

\begin{figure}[t!]
\begin{center}
\includegraphics[width=0.85\textwidth]{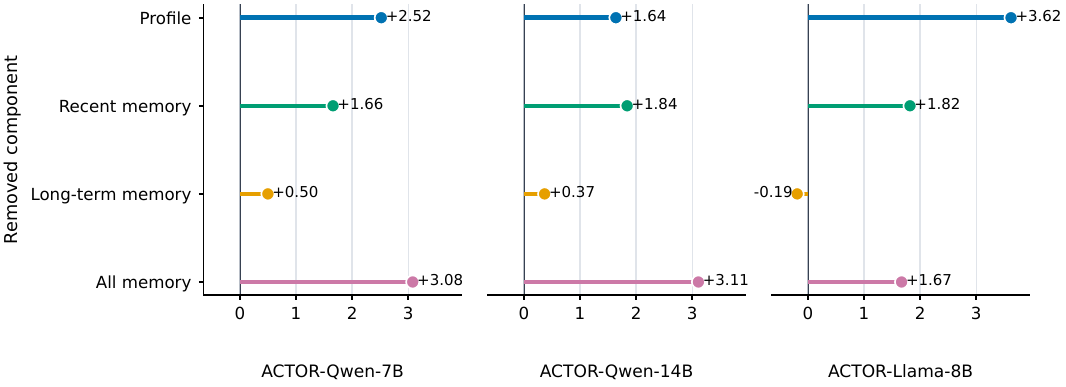} 
\end{center}
\caption{Accuracy changes after removing different context components.
Each panel corresponds to one ACTOR model, and the vertical reference line
denotes the full-context performance. Positive values indicate performance
drops after removing the component.} 
\label{fig:memory_ablation} 
\end{figure}

Figure~\ref{fig:memory_ablation} reports the context ablation under the
anonymous setting. We keep the Current Scene unchanged and separately remove
the Character Profile, Recent Memory, or Long-term Memory from the input
context. We find that \textbf{removing either the Character Profile or Recent Memory consistently reduces the performance of all three ACTOR models.} This indicates stable character traits and recent behavioral context provide complementary signals for global personality reasoning and local behavioral adaptation. In contrast, removing Long-term Memory changes the accuracy of the two Qwen models by no more than 0.5 percentage points and slightly improves the Llama model. This difference may reflect the partial overlap between Long-term Memory and the Character Profile, making the marginal contribution of long-term information dependent on the backbone.

\begin{figure}[t!]
\begin{center}
\includegraphics[width=0.85\textwidth]{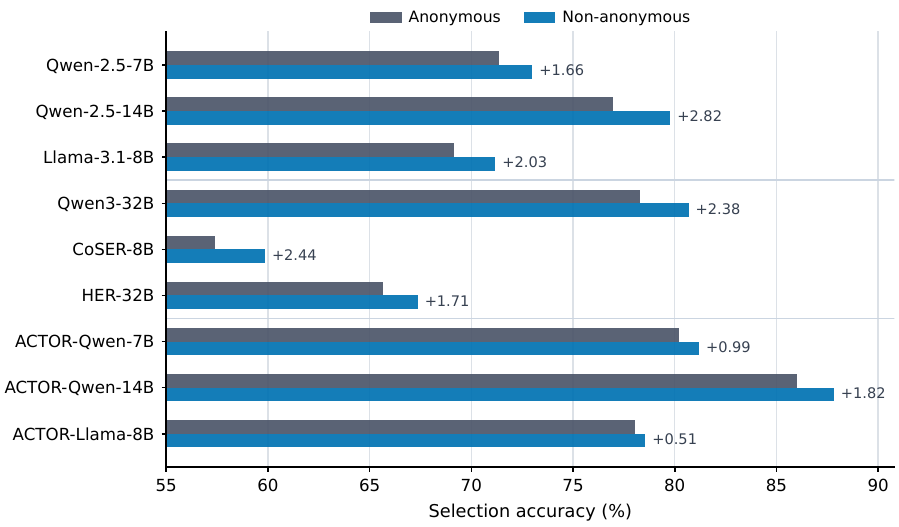} 
\end{center}
\caption{Selection accuracy under the anonymous and identity-provided settings.} 
\label{fig:identity_ablation} 
\end{figure}

Figure~\ref{fig:identity_ablation} compares model performance under anonymous and identity-provided settings. Supplying book titles and character names boosts accuracy for all models, as explicit identities help retrieve prior knowledge of familiar characters. ACTOR’s gains (0.51–1.82 pp) are smaller than those of baselines. \textbf{ACTOR's advantage therefore does not primarily depend on explicit
character identities.} Its strong anonymous-setting performance shows the model leverages provided character information and scene context, rather than memorized knowledge of known books or characters. This benefit is valuable for role-playing original or unseen characters.

\section{Conclusion}

We presented \textbf{ACTOR}, an automated framework for extracting and chronologically organizing dialogue and persona-relevant non-dialogue behaviors from literary texts. By pairing source behaviors with counterfactual alternatives, ACTOR formulates literary role-playing as an action-centered temporal prediction task and trains models with an RPO objective that combines preference learning with direct supervision on source behaviors. Experiments show that ACTOR consistently improves next-behavior prediction across different settings, with gains on unseen works indicating transfer beyond memorizing training trajectories. Ablation and qualitative analyses further highlight the importance of updated character profiles and recent behavioral memory. Supplementary evaluation with the CoSER evaluation framework suggests that these benefits extend to open-ended character interaction, although the framework may inherit noise from automatic data construction and currently relies on separate selection and generation protocols. Future work will improve construction reliability, broaden language and genre coverage, and develop unified evaluations.

\section*{AI Use Statement}

We used generative AI tools to assist with extracting and cleaning character behavior data from literary texts, including behavior structuring, candidate key-behavior identification, temporal information organization, and counterfactual behavior generation. Generative AI was also used to assist with drafting and polishing the manuscript and with generating and revising figures. All AI-generated data were subject to programmatic validation, source-based evidence verification, and manual inspection. The authors also reviewed all AI-assisted text and figures. Generative AI was not used to replace the authors in formulating the core research questions, drawing theoretical conclusions, or interpreting experimental results. The authors take full responsibility for the final content, research claims, and all artifacts produced with the assistance of generative AI.

\section*{Ethics statement}

This work involves no human participants, private personal data, or direct interaction with real individuals, so institutional review board approval is unnecessary. We construct character behavior trajectories from literary texts. To comply with copyright and licensing constraints, our released resource contains only derived anonymized records. Full original novels are not redistributed.

Automatically generated behaviors and counterfactuals may carry biases inherited from source literature. Biases can also be introduced by language models. Such biases may produce stereotyped portrayals. These portrayals touch on gender, culture, and social roles. For this reason, the resulting models cannot serve as reliable depictions of real people. They must not be deployed for high-stakes decision-making. We mitigate annotation errors through source-based validation alongside structural auditing. Manual inspection further supports this process. We take responsibility for ethical usage and thorough documentation of all released materials.

\section*{Reproducibility statement}

The ACTOR data-construction pipeline is described in Section~\ref{sec:dataset_construct}. The RPO training objective and its model-specific settings are provided in Section~\ref{action_training}. The
data splits, evaluated models, task formulation, and open-ended evaluation
protocol are specified in Section~\ref{sec:evaluation_setup}. The appendix further provides dataset
statistics and structural audits (Appendix~\ref{data_statistics}), the complete dataset schema (Appendix~\ref{app:dataset_schema}) and the prompts used (Appendix~\ref{app:prompt}). Code and data are available at https://github.com/SilentStar233/ACTOR.

\bibliography{iclr2027_conference}
\bibliographystyle{iclr2027_conference}

\appendix
\section{Dataset Statistics}
\label{data_statistics}

The final anonymized corpus contains behavior records from 143 non-empty literary works. The training partition contains 270,856 source preference records from 115 works. The evaluation partition contains 17,560 four-way next-behavior questions from 143 works: 11,770 temporal-evaluation instances from the 115 training works and 5,790 cross-work instances from 28 additional works. The evaluation set contains 8,221 key-behavior instances and 9,339 non-key-behavior instances. Table~\ref{tab:appendix_corpus_summary} summarizes the corpus composition.

\begin{table*}[t]
\caption{Summary of the dataset.}
\label{tab:appendix_corpus_summary}
\centering
\resizebox{\textwidth}{!}{%
\begin{tabular}{lrrrrrr}
\toprule
Split & Works & Records & Characters & Scenes & Key & Non-key \\
\midrule
Training & 115 & 270,856 & 3,939 & 30,818 & 26,374 & 244,482 \\
Evaluation & 143 & 17,560 & 2,437 & 11,201 & 8,221 & 9,339 \\
Temporal / cross-work & 115 / 28 & 17,560 & -- & -- & -- & -- \\
\bottomrule
\end{tabular}%
}
\end{table*}

\begin{figure}[t!]
\begin{center}
\includegraphics[width=1.0\textwidth]{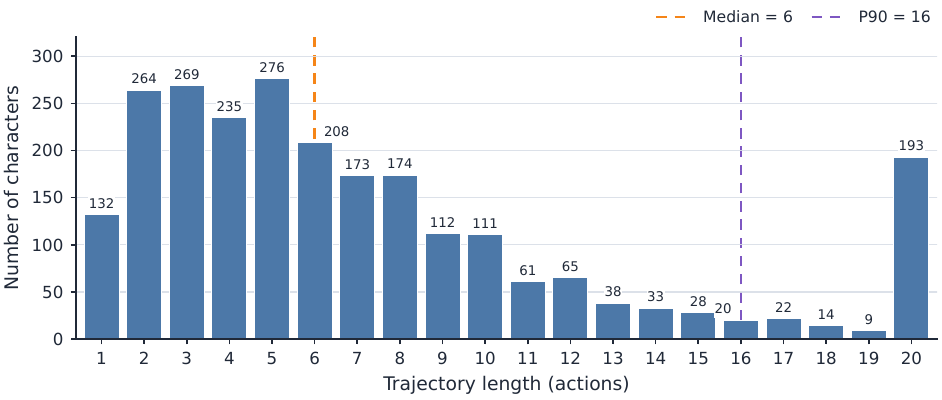} 
\end{center}
\caption{Evaluation trajectory-length distribution.} 
\label{fig:eval_trajectory_length} 
\end{figure}

\begin{figure}[t!]
\begin{center}
\includegraphics[width=1.0\textwidth]{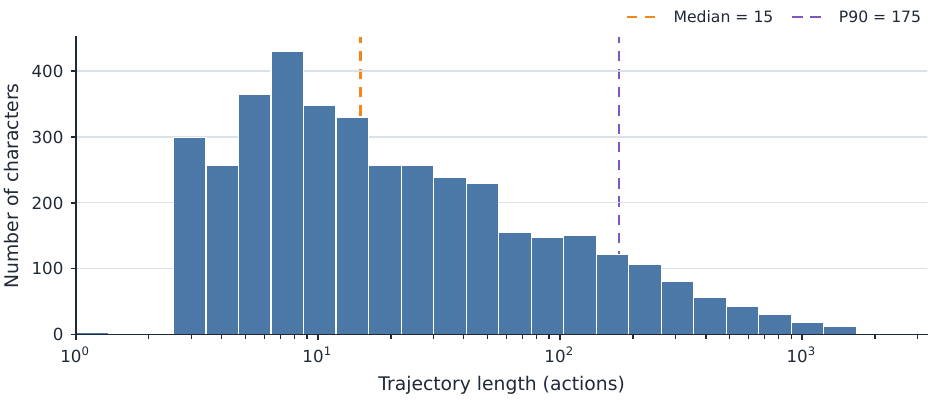} 
\end{center}
\caption{Training trajectory-length distribution.} 
\label{fig:train_trajectory_length} 
\end{figure}

The context supplied to each instance is compact but structured. In the evaluation set, the median character trajectory contains 6 recorded behaviors, the median profile contains 120 whitespace-delimited words, and the median recent and long-term memories contain 8 and 7 entries, respectively. Training records have longer trajectories because they include the complete training prefix; their median trajectory length is 15 behaviors. Profile freshness is reported per record and indicates whether the profile was immediately updated, moderately distant, or stale relative to the target behavior. Trajectory-length distributions are shown in Figures~\ref{fig:eval_trajectory_length} and~\ref{fig:train_trajectory_length}.

We also audit the structural validity of the released records. The rebuilding validation retained 288,416 records and excluded 279 records before release. All 17,560 evaluation instances contain exactly four options and a valid answer, while all 270,856 training records contain non-empty chosen and rejected behaviors. No duplicate record identifiers or malformed placeholder strings were detected. A small number of participant-description keys are composite or relationship-style expressions rather than single ``Character $n$'' placeholders; they are reported separately for manual review and are not counted as known-alias or malformed-placeholder failures. The complete audit is given in Table~\ref{tab:appendix_quality_audit}.

\begin{table*}[t]
\caption{Evaluation-question and preference-pair audit.}
\label{tab:appendix_quality_audit}
\centering
\resizebox{\textwidth}{!}{%
\begin{tabular}{lrrr}
\toprule
Audit & Training & Evaluation & Result \\
\midrule
Records with required answer fields & 270,856 & 17,560 & valid \\
Invalid preference/options records & 0 & 0 & none \\
Duplicate record IDs & 0 & 0 & none \\
Malformed placeholders & 0 & 0 & none \\
Non-placeholder participant-key occurrences & 764 & 65 & manual review \\
Four-option questions & -- & 17,560 & all evaluation records \\
\bottomrule
\end{tabular}%
}
\end{table*}

\begin{figure}[t!]
\begin{center}
\includegraphics[width=1.0\textwidth]{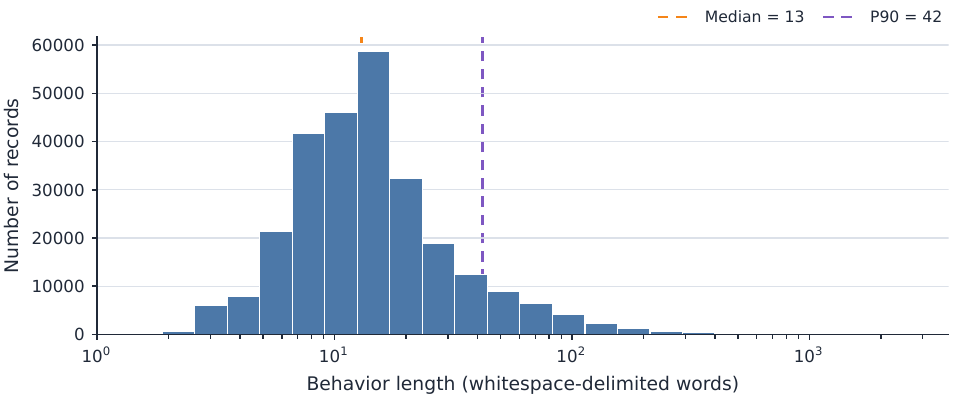} 
\end{center}
\caption{Length distribution of chosen behaviors.} 
\label{fig:chosen_length_words} 
\end{figure}

\begin{figure}[t!]
\begin{center}
\includegraphics[width=1.0\textwidth]{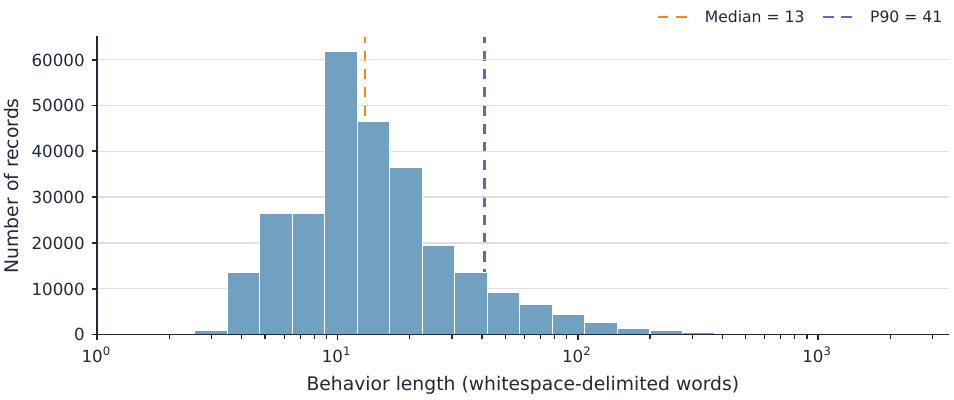} 
\end{center}
\caption{Length distribution of rejected behaviors.} 
\label{fig:rejected_length_words} 
\end{figure}

Finally, source chosen and rejected behaviors are closely matched in surface length: their medians are both 13 whitespace-delimited words, with means of 21.69 and 21.30 words, respectively. The corresponding ranges and percentiles are reported in Table~\ref{tab:appendix_preference_lengths}; the distributions are visualized in Figures~\ref{fig:chosen_length_words} and~\ref{fig:rejected_length_words}. 

\begin{table*}[t]
\caption{Length statistics for source preference pairs, measured in whitespace-delimited words.}
\label{tab:appendix_preference_lengths}
\centering
\resizebox{\textwidth}{!}{%
\begin{tabular}{lrrrrrr}
\toprule
Response & Count & Mean & Median & P10 & P90 & Min--Max \\
\midrule
Chosen & 270,856 & 21.6877 & 13 & 6 & 42 & 1--3635 \\
Rejected & 270,856 & 21.3041 & 13 & 5 & 41 & 2--3297 \\
\bottomrule
\end{tabular}%
}
\end{table*}

\section{Dataset Schema}
\label{app:dataset_schema}

Each ACTOR instance is centered on a target character and a behavior occurring at a particular point in the story. All instances contain the character profile, behavioral memories, and current-scene context available before the target behavior. Training instances additionally contain a preferred source behavior and a rejected counterfactual, whereas evaluation instances contain four candidate behaviors and the corresponding answer label. All textual fields are anonymized before being provided to the model.

\begin{table*}[t]
\caption{Context fields shared by ACTOR training and evaluation instances.}
\label{tab:dataset_context_schema}
\centering
\renewcommand{\arraystretch}{1.35}
\setlength{\tabcolsep}{7pt}
\resizebox{\textwidth}{!}{%
\begin{tabular}{p{0.23\textwidth}p{0.17\textwidth}p{0.52\textwidth}}
\toprule
Field & Type & Description \\
\midrule
\emph{character}
& Anonymized label
& Anonymized name of the target character shown to the model. \\

\emph{profile}
& Structured object
& The latest character-profile version available before the target behavior. It summarizes persistent traits, relationships, and established states. \\

\emph{long\_term\_memory}
& List of T-A-O records
& Previously verified key behaviors that provide long-range behavioral history. \\

\emph{recent\_memory}
& List of T-A-O records
& Behaviors from the most recent preceding scenes, representing the local behavioral context. \\

\emph{scene}
& Structured object
& The current-scene description, temporal information, present participants, and anonymized participant descriptions. It excludes the target behavior and all subsequent events. \\

\emph{is\_critical}
& Boolean
& Indicates whether the target behavior is a verified key behavior. \\

\emph{action\_uid}, \emph{scene\_uid}
& Identifiers
& Internal identifiers for the target behavior and its containing scene. They are used for traceability and are not required as model inputs. \\
\bottomrule
\end{tabular}%
}
\end{table*}

\begin{table*}[t]
\caption{Split-specific fields in ACTOR training and evaluation instances.}
\label{tab:dataset_target_schema}
\centering
\renewcommand{\arraystretch}{1.35}
\setlength{\tabcolsep}{7pt}
\resizebox{\textwidth}{!}{%
\begin{tabular}{p{0.23\textwidth}p{0.17\textwidth}p{0.52\textwidth}}
\toprule
Field & Split & Description \\
\midrule
\emph{chosen}
& Training
& The behavior performed by the target character in the source text. It is treated as the preferred response during RPO training. \\

\emph{rejected}
& Training
& A contextually plausible but character-inconsistent counterfactual behavior paired with the source behavior. \\

\emph{options}
& Evaluation
& Four candidate next behaviors, consisting of one source behavior and three counterfactual behaviors. \\

\emph{correct\_answer}
& Evaluation
& The identifier of the candidate corresponding to the source behavior. \\

\emph{character\_action\_order}
& Training / Evaluation
& The position of the target behavior in the character-level chronological trajectory. \\

\emph{split}
& Training
& Partition label indicating that the instance belongs to the training prefix of a character trajectory. \\

\emph{sample\_id}, \emph{question\_id}
& Training / Evaluation
& Unique identifiers used to trace each instance during construction, validation, and evaluation. \\
\bottomrule
\end{tabular}%
}
\end{table*}

Table~\ref{tab:dataset_example} presents an actual anonymized evaluation
instance from the released dataset. The textual fields and candidate options
are preserved as stored in the dataset; only internal bookkeeping fields are
omitted from the table.

\begin{table*}[t]
\caption{An actual anonymized next-behavior selection instance from the ACTOR
evaluation set.}
\label{tab:dataset_example}
\centering
\renewcommand{\arraystretch}{1.45}
\setlength{\tabcolsep}{8pt}
\resizebox{\textwidth}{!}{%
\begin{tabular}{p{0.22\textwidth}p{0.70\textwidth}}
\toprule
Component & Dataset content \\
\midrule
Target character
& Character 1 \\

Character profile
& Character 1 is a young law student and the heir presumptive to Character 2,
positioned by expectation as the future baronet. Beyond this social and
familial standing, no stable temperament, values, interaction style, personal
goals, or relationship orientation are evidenced before his first recorded
action. \\

Recent memory
& \textbf{Thought:} I do not feel eager to be drawn into the family's
notice. \newline
\textbf{Action:} Receives Character 2 overtures without warmth. \newline
\textbf{Observation:} Character 2 persists, making allowance for what he treats
as youthful modesty. \\

Long-term memory
& \textbf{Thought:} I do not feel eager to be drawn into the family's
notice. \newline
\textbf{Action:} Receives Character 2 overtures without warmth. \newline
\textbf{Observation:} Character 2 persists, making allowance for what he treats
as youthful modesty. \\

Current scene
& \textbf{Time:} in one of their spring excursions to London, when Character 3
was in her first bloom. \newline
\textbf{Present:} Character 3, Character 2, Character 1. \newline
The scene has just begun. Character 3 is the eldest daughter of Character 2
and Character 4; Character 2 is a baronet and the father of Character 3,
Character 5, and Character 6; Character 1 is the heir presumptive to
Character 2 and a young law student expected to become the future baronet. \\

Candidate A
& Behaves agreeably during the introduction. \\

Candidate B
& Regards Character 3 with open disdain throughout. \\

Candidate C
& Defers slavishly to Character 2 guidance. \\

Candidate D
& Presses Character 3 hand with ardent presumption. \\

Correct answer
& A \\
\bottomrule
\end{tabular}%
}
\end{table*}

For training instances, the same contextual fields are paired with \emph{chosen} and \emph{rejected} behaviors instead of four candidate options. This shared schema enables the extracted trajectories to support both preference-based training and objective next-behavior evaluation.

\section{Detailed Case-Study Instances}
\label{app:case_details}

Tables~\ref{tab:case_detail_state}--\ref{tab:case_detail_action} provide
additional details for the three cases discussed in
Section~\ref{sec:case_study}. To maintain readability, we retain the evidence
most relevant to each behavioral decision and use ellipses to indicate omitted
profile, memory, or scene content.

\begin{table*}[t]
\caption{Detailed instance for Case 1, which requires tracking a long-term
character transformation. Ellipses indicate omitted context.}
\label{tab:case_detail_state}
\centering
\renewcommand{\arraystretch}{1.12}
\resizebox{0.95\textwidth}{!}{%
\begin{tabular}{p{0.16\textwidth}p{0.78\textwidth}}
\hline
\textbf{Field} & \textbf{Content} \\
\hline

Profile
&
Character 1 was formerly cold, solitary, suspicious, anti-Christmas, and
dismissive of poverty. After witnessing the consequences of his former life,
he becomes grateful, humble, playful, openly generous, and eager to repair his
past hardness. \ldots
\\
\hline

Long-term memory
&
\textbf{Thought:} I now see that this neglected and plundered dead man's case
may be my own if my present life continues in this direction.
\textbf{Action:} He recognizes that his own life is tending toward the same
unloved death.
\textbf{Observation:} \ldots

\textbf{Thought:} I must bind myself to a changed life and preserve the
lessons I have learned.
\textbf{Action:} He vows to honor Christmas throughout the year and to live
by the lessons of the past, present, and future.
\textbf{Observation:} \ldots
\\
\hline

Recent memory
&
He goes to church, walks through the streets with pleasure, treats children
kindly, and questions beggars rather than ignoring them. He then approaches
his nephew's house, overcomes his anxiety, and joins the family gathering.
\ldots
\\
\hline

Current scene
&
The character waits for his clerk, who arrives eighteen and a half minutes
late. He initially challenges the clerk in a severe tone and pretends that he
will no longer tolerate such conduct. His severity is described as feigned and
suddenly playful, although the clerk remains alarmed.
\\
\hline

Candidates
&
\textbf{A:} Punishes the clerk for arriving late and refuses to provide
additional warmth or assistance.

\textbf{B:} Offers only conditional assistance after scrutinizing the clerk's
family and conduct.

\textbf{C:} Wishes the clerk a merry Christmas, raises his salary, promises to
assist his family, and asks him to add coal to the fire.

\textbf{D:} Offers help primarily as a public demonstration of his own
generosity.
\\
\hline

Predictions
&
Source behavior: \textbf{C};
Qwen-2.5-Instruct 7B: \textbf{A};
ACTOR-Qwen-2.5-7B: \textbf{C}.
\\
\hline
\end{tabular}%
}
\end{table*}

\begin{table*}[t]
\caption{Detailed instance for Case 2, which requires maintaining local
dialogue continuity. Ellipses indicate omitted context.}
\label{tab:case_detail_dialogue}
\centering
\renewcommand{\arraystretch}{1.12}
\resizebox{0.95\textwidth}{!}{%
\begin{tabular}{p{0.16\textwidth}p{0.78\textwidth}}
\hline
\textbf{Field} & \textbf{Content} \\
\hline

Profile
&
Character 1 is identified only as a chambermaid responsible for room care.
No personal traits, social relationships, motives, history, or attitudes are
available in the profile.
\\
\hline

Long-term memory
&
The chambermaid has previously revealed that Character 3 visited nearly every
day and that the visits were made to Character 4. She further reveals that
Character 4 and Character 3 sometimes went out together. \ldots
\\
\hline

Recent memory
&
When asked about Character 4, the chambermaid answers directly that Character 4
went out only a few minutes earlier. Character 2 finds the news strange and
hurriedly searches through his belongings.
\\
\hline

Current scene
&
During an extended conversation, the chambermaid alternates between direct
answers, teasing remarks, and evasive responses. She discusses Character 3's
visits, tells Character 2 that the visits occurred nearly every day, and
reveals that Character 3 and Character 4 sometimes went out together.
Character 2 becomes visibly disturbed and questions her about the details.
\ldots
\\
\hline

Candidates
&
\textbf{A:} promises: ``Certainly not.''

\textbf{B:} greedily hints: ``Pay me.''

\textbf{C:} haughtily says: ``Never, sir.''

\textbf{D:} eagerly says: ``I'll tell.''
\\
\hline

Predictions
&
Source behavior: \textbf{A};
Qwen-2.5-Instruct 7B: \textbf{D};
ACTOR-Qwen-2.5-7B: \textbf{A}.
\\
\hline
\end{tabular}%
}
\end{table*}

\begin{table*}[t]
\caption{Detailed instance for Case 3, which requires adapting an ongoing
action to a change in the current scene. Ellipses indicate omitted context.}
\label{tab:case_detail_action}
\centering
\renewcommand{\arraystretch}{1.12}
\resizebox{0.95\textwidth}{!}{%
\begin{tabular}{p{0.16\textwidth}p{0.78\textwidth}}
\hline
\textbf{Field} & \textbf{Content} \\
\hline

Profile
&
Character 1 is a steward and trusted envoy whose loyalty lies primarily with
his mistress and her political faction. Although normally self-protective
under direct threat, he is willing to use violence against Character 5 to
demonstrate his factional loyalty. \ldots
\\
\hline

Long-term memory
&
\textbf{Thought:} If I meet Character 5, killing him will prove my allegiance
to the party I serve.
\textbf{Action:} He declares that, if he encounters Character 5, he will show
which party he follows.
\textbf{Observation:} \ldots

\textbf{Thought:} Here is the proclaimed traitor I hoped to meet; killing him
can raise my fortunes.
\textbf{Action:} He draws his sword and threatens to kill Character 5.
\textbf{Observation:} Character 10 intervenes between them.
\\
\hline

Recent memory
&
The character repeatedly follows his mistress's instructions, refuses to
disclose her confidential letter, and prioritizes her orders despite warnings
that the journey is dangerous. \ldots
\\
\hline

Current scene
&
Character 1 prepares to attack Character 5. Character 10 steps between them
and refuses to move. Character 1 threatens Character 10 and orders him to
stand aside, but Character 10 remains defiant and prepares to resist the
attack.
\\
\hline

Candidates
&
\textbf{A:} Kneels before Character 10.

\textbf{B:} Casts away his sword.

\textbf{C:} Fights Character 10.

\textbf{D:} Shields Character 5, whom he had intended to kill.
\\
\hline

Predictions
&
Source behavior: \textbf{C};
Qwen-2.5-Instruct 7B: \textbf{D};
ACTOR-Qwen-2.5-7B: \textbf{C}.
\\
\hline
\end{tabular}%
}
\end{table*}

\section{CoSER Interaction Examples}
\label{app:coser_examples}

Tables~\ref{tab:coser-case-michelangelo} and~\ref{tab:coser-case-heathcliff} present two qualitative comparisons from the CoSER evaluation. Each model is evaluated in a separate simulation of the same initial scene. Context and character descriptions are summarized from the evaluation inputs. We select excerpts from the first eight character messages in each simulation, omitting next-speaker predictions and environment messages. Turn numbers preserve their original order; gaps indicate omitted turns, and ellipses mark omissions within a turn. Speaker attributions follow the recorded role fields, including inconsistencies; names are shortened in the first table and redundant name prefixes are omitted. Scores are scene-level averages of the four CoSER dimensions, computed from the full simulations rather than these excerpts. These selected cases illustrate interaction patterns and limitations.

\begin{table*}[t]
\caption{Conversational progression in \emph{The Agony and the Ecstasy} (scene 0006). The scene context is summarized and model outputs are excerpted from the recorded simulations.}
\label{tab:coser-case-michelangelo}
\centering
\begingroup
\small
\setlength{\tabcolsep}{0pt}
\renewcommand{\arraystretch}{1.12}
\resizebox{\textwidth}{!}{%
\begin{tabular}{@{}p{\textwidth}@{}}
\toprule
\textbf{Scene.} Michelangelo and Tommaso examine a wooden model of St. Peter's dome. After years of design work, Michelangelo is conscious of his declining health and wants his vision to survive him. \\[4pt]
\midrule
\textbf{Characters and motivations.} Michelangelo is independent and devoted to his craft; he wants Tommaso to understand the design. Tommaso admires his work, worries about his health, and wants to support him. \\[4pt]
\midrule
\textbf{Qwen-2.5-Instruct 7B}\hfill\textbf{Scene score: 26.75}\par\smallskip
  \emph{1.} \textbf{Michelangelo:} \ldots{} I stand before the model of the dome, my hands tracing the lines with a steady hand. \ldots{}\par\smallskip
  \emph{2.} \textbf{Tommaso:} \ldots{} (Tommaso, you're my inspiration. Let's do this together.)\par\smallskip
  \emph{3.} \textbf{Michelangelo:} \ldots{} Together, we shall see this through. (We shake hands, a silent pact between us.)\par\smallskip
  \emph{4.} \textbf{Tommaso:} (Our hands meet, a bond of friendship and shared purpose.) \ldots{}\par\smallskip
  \emph{5.} \textbf{Michelangelo:} (Our hands meet, a bond of friendship and shared purpose.) (I return his smile, a flicker of hope lighting my eyes.) \ldots{}\par\smallskip
  \emph{6.} \textbf{Tommaso:} (Our hands meet, a bond of friendship and shared purpose.) (I return his smile, a flicker of hope lighting my eyes.) \ldots{} \\[4pt]
\midrule
\textbf{CoSER-8B}\hfill\textbf{Scene score: 19.50}\par\smallskip
  \emph{3.} \textbf{Michelangelo:} \ldots{} What do you think of making it more concave?\par\smallskip
  \emph{4.} \textbf{Tommaso:} \ldots{} What inspired this change?\par\smallskip
  \emph{5.} \textbf{Michelangelo:} \ldots{} I just feel that it needs more... movement. But I fear I may be losing my perspective.\par\smallskip
  \emph{6.} \textbf{Tommaso:} \ldots{} Your vision is unparalleled. Trust in your instincts. \ldots{}\par\smallskip
  \emph{8.} \textbf{Tommaso:} \ldots{} Your work will outlive us all. How can I assist you in bringing this vision to life? \\[4pt]
\midrule
\textbf{ACTOR-Qwen-2.5-7B}\hfill\textbf{Scene score: 46.00}\par\smallskip
  \emph{1.} \textbf{Michelangelo:} \ldots{} "Tommaso, I have a new idea for the dome," \ldots{}\par\smallskip
  \emph{2.} \textbf{Tommaso:} \ldots{} "But why not use a flying buttress?" \ldots{}\par\smallskip
  \emph{3.} \textbf{Michelangelo:} \ldots{} "It would support the weight, but it would also be visible. I want the dome to be a perfect sphere, hidden from view."\par\smallskip
  \emph{5.} \textbf{Michelangelo:} \ldots{} "It's not just a building, Tommaso, it's a symbol."\par\smallskip
  \emph{6.} \textbf{Tommaso:} \ldots{} "A symbol of what, Michelangelo?"\par\smallskip
  \emph{7.} \textbf{Michelangelo:} \ldots{} "Of God's creation, of man's ingenuity, of our faith and our dreams." \\[4pt]
\bottomrule
\end{tabular}%
}
\endgroup
\end{table*}

In Table~\ref{tab:coser-case-michelangelo}, the Qwen-2.5-Instruct 7B backbone initially attends to the dome model, but Tommaso's next turn addresses himself by name. The interaction then settles into declarations of shared purpose, and both characters repeat almost the same handshake description instead of discussing the design. CoSER-8B avoids that repetition and raises a question about the dome's shape, although the exchange soon turns to general reassurance. ACTOR-Qwen-2.5-7B sustains the subject more clearly: Tommaso asks about a flying buttress, Michelangelo responds in terms of support and visibility, and later turns discuss what the dome symbolizes. \textbf{The useful contrast is whether the conversation develops beyond the initial scene cue.} The backbone recognizes that cue but does not maintain a distinct or progressing exchange; ACTOR does so in these selected turns. This reading concerns conversational progression, not the architectural or historical correctness of the design claims.

\begin{table*}[t]
\caption{Scene alignment and speaker attribution in \emph{Wuthering Heights} (scene 0189). The scene context is summarized and model outputs are excerpted from the recorded simulations.}
\label{tab:coser-case-heathcliff}
\centering
\begingroup
\small
\setlength{\tabcolsep}{0pt}
\renewcommand{\arraystretch}{1.12}
\resizebox{\textwidth}{!}{%
\begin{tabular}{@{}p{\textwidth}@{}}
\toprule
\textbf{Scene.} During Heathcliff's final days, Ellen Dean visits his chamber. He has withdrawn from company and refuses food, while she worries about his deteriorating condition. \\[4pt]
\midrule
\textbf{Characters and motivations.} Ellen Dean is practical and compassionate, with a strong sense of duty toward those in her care. Heathcliff is preoccupied with rejoining Catherine and resists interruption. Her concern conflicts with his withdrawal. \\[4pt]
\midrule
\textbf{Qwen-2.5-Instruct 7B}\hfill\textbf{Scene score: 23.75}\par\smallskip
  \emph{1.} \textbf{Ellen Dean:} \ldots{} (I place the tray on the table, my hand trembling slightly as I watch Heathcliff's shallow breathing.)\par\smallskip
  \emph{2.} \textbf{Heathcliff:} [Heathcliff's fate is sealed, yet I must fulfill my duty.] (I approach the bed, my steps hesitant, and gently place the tray beside him, hoping he might find some solace in the food.)\par\smallskip
  \emph{3.} \textbf{Ellen Dean:} \ldots{} "Eat, Mr. Heathcliff. It's the least I can do."\par\smallskip
  \emph{4.} \textbf{Heathcliff:} \ldots{} "Eat, Mr. Heathcliff. It's the least I can do."\par\smallskip
  \emph{7.} \textbf{Heathcliff:} \ldots{} The wind howls mournfully, as if echoing his fate. \ldots{} \\[4pt]
\midrule
\textbf{CoSER-8B}\hfill\textbf{Scene score: 26.50}\par\smallskip
  \emph{2.} \textbf{Heathcliff:} \ldots{} I am... better. Thank you.\par\smallskip
  \emph{3.} \textbf{Ellen Dean:} \ldots{} I've brought you some food, sir. Shall I open the windows?\par\smallskip
  \emph{6.} \textbf{Heathcliff:} \ldots{} Do you... do you miss me, Nelly?\par\smallskip
  \emph{7.} \textbf{Ellen Dean:} \ldots{} I... I do, Mr. Heathcliff. You've been kind to me in the past.\par\smallskip
  \emph{8.} \textbf{Heathcliff:} \ldots{} Thank you, Nelly. That's all I need to know. \\[4pt]
\midrule
\textbf{ACTOR-Qwen-2.5-7B}\hfill\textbf{Scene score: 47.50}\par\smallskip
  \emph{1.} \textbf{Ellen Dean:} \ldots{} Heathcliff, you've not eaten. I brought you some broth.\par\smallskip
  \emph{2.} \textbf{Heathcliff:} [Heathcliff's eyes flicker with a mix of anger and despair as he looks at the food.]\par\smallskip
  \emph{3.} \textbf{Ellen Dean:} (Heathcliff looks up, his eyes hollow and distant) Nelly, I’ve no appetite. (His voice is a hoarse whisper) I’ve no need for food.\par\smallskip
  \emph{5.} \textbf{Ellen Dean:} \ldots{} I’ve seen enough of this world. (His voice is a mere shadow of its former self) I’ve tasted enough of its bitterness.\par\smallskip
  \emph{8.} \textbf{Ellen Dean:} \ldots{} Heathcliff, you’re not yourself. (She tries to keep her voice steady) You’ve been ill for days. \\[4pt]
\bottomrule
\end{tabular}%
}
\endgroup
\end{table*}

Table~\ref{tab:coser-case-heathcliff} shows a different failure in the Qwen-2.5-Instruct 7B backbone. It introduces Ellen's concern and the tray of food, but the turn labelled Heathcliff adopts the caregiver's perspective and then repeats Ellen's appeal to eat almost verbatim. Later turns dwell on the wind and the room rather than developing a response to his refusal. CoSER-8B maintains a reciprocal dialogue, yet shifts from the supplied conflict toward a reassuring exchange. ACTOR-Qwen-2.5-7B keeps the offered food, refusal to eat, and withdrawal in view, but turns 3 and 5 are labelled Ellen Dean while voicing Heathcliff's perspective. \textbf{Stronger engagement with the scene therefore does not resolve speaker attribution.} Although ACTOR has the highest average score here, its character-fidelity score is lower than the backbone's (42.50 versus 62.50); the scene average should not be interpreted as improvement on every dimension.

Together, the examples suggest that ACTOR's behavioral supervision can support more sustained exchanges and closer engagement with a scene's conflict in open-ended interaction. They also expose a remaining failure mode: maintaining the correct speaker and perspective across turns. The examples complement the aggregate CoSER results without establishing how often either pattern occurs.

\section{prompt}
\label{app:prompt}

To facilitate reproducibility, this section lists the prompts used throughout the ACTOR model. Tables~\ref{tab:prompt_plot_context}--\ref{tab:prompt_auditing} present the prompts adopted in the action trajectory extraction stage. Tables~\ref{tab:prompt_profile}--\ref{tab:prompt_counterfactual} are for dynamic character-profile construction and counterfactual rewriting. Tables~\ref{tab:prompt_training}--\ref{tab:prompt_evaluation} provide the shared context and output prompts used for RPO training and four-way next-behavior evaluation.

\begin{table*}[t]
\caption{Prompts for plot and character-context curation.}
\label{tab:prompt_plot_context}
\centering
\resizebox{\textwidth}{!}{%
\begin{tabular}{p{0.20\textwidth}p{0.74\textwidth}}
\toprule
\textbf{Prompt} & \textbf{Template and requirements} \\
\midrule

\textbf{Plot Segmentation}
&
\textit{(intro)} Partition the supplied book chunk into consecutive,
non-overlapping plot segments for downstream character-action extraction.

\textit{(requirements)} Return the exact first and last sentences of every
segment. Preserve source-text order and continue truncated segments from the
previous chunk. Maintain the canonical character registry and report newly
observed aliases. Do not create overlapping segments or include a broad parent
plot together with a contained sub-plot. Return strict JSON only.
\\
\midrule

\textbf{Overlap Repair}
&
\textit{(intro)} Repartition a source-text region that has been assigned to
overlapping plot segments.

\textit{(requirements)} Cover the supplied region exactly once, without gaps or
duplicated passages. Segments must be mutually exclusive and remain in source
order. Start a new segment only when there is a meaningful change in the
immediate goal, setting, interaction, or outcome. Return a parsimonious set of
segments with exact first-sentence anchors.
\\
\midrule

\textbf{Character Enrichment}
&
\textit{(intro)} Generate compact character details for each supplied character
in a finalized plot.

\textit{(requirements)} Use only the plot summary and the supplied plot text.
For every character, generate a short description of identity, social position,
or relevant relationships, together with a concise account of the character's
actions, decisions, intentions, or experiences in this plot. Do not use outside
knowledge or information from later plots. Preserve the supplied canonical
names and return strict JSON.
\\
\midrule

\textbf{Alias Grouping}
&
\textit{(intro)} Group character names that refer to the same person and select
a canonical name.

\textit{(requirements)} Merge formal names, honorifics, nicknames, and
diminutives only when the identity is supported. Do not merge family members
merely because they share a surname or title. If a name is genuinely ambiguous,
place it in the unresolved list rather than guessing. Every input name must
appear exactly once in either a group or the unresolved list.
\\

\bottomrule
\end{tabular}%
}
\end{table*}

\begin{table*}[t]
\caption{Prompts for scene and T-A-O behavior extraction.}
\label{tab:prompt_behavior_extraction}
\centering
\resizebox{\textwidth}{!}{%
\begin{tabular}{p{0.20\textwidth}p{0.74\textwidth}}
\toprule
\textbf{Prompt} & \textbf{Template and requirements} \\
\midrule

\textbf{Scene Splitting}
&
\textit{(intro)} Split a finalized plot into consecutive scenes representing
continuous interactions in one setting.

\textit{(requirements)} Scenes must jointly cover the entire plot text without
gaps or overlaps. Start a new scene when the setting changes or the interaction
shifts to a different focus. For each scene, provide a description, canonical
participants, relationship context, and the exact first sentence used to locate
the scene in the source text.
\\
\midrule

\textbf{T-A-O Extraction}
&
\textit{(intro)} Extract character behavior trajectories as
Thought--Action--Observation loops from the supplied plot text.

\textit{(requirements)} An action is a decision-bearing behavior for which the
character has plausible alternatives. Include consequential decisions,
decisional speech such as refusal, persuasion, promise, deception, or threat,
and deliberate silence or non-intervention. Exclude passive information
exchange, routine movements, decorative gestures, and actions with no meaningful
character choice.

Each step must contain exactly one atomic decision. Thought is written from the
character's first-person perspective and must be grounded in the source text.
Action describes the performed behavior or preserves a direct quotation when
applicable. Observation records only reactions or environmental changes
perceptible to the character; omniscient narration and later events are
prohibited.

For every behavior, generate exactly three situationally plausible but
character-inconsistent alternatives. Mark a behavior as critical only when the
character has a reasonable alternative and the selected behavior changes a
persistent state or a future situational possibility. Record the scene's actual
time and supporting evidence, rather than dates mentioned only in plans,
memories, or documents. Return strict JSON only.
\\

\bottomrule
\end{tabular}%
}
\end{table*}

\begin{table*}[t]
\caption{Prompts for criticality verification and timeline reconstruction.}
\label{tab:prompt_auditing}
\centering
\resizebox{\textwidth}{!}{%
\begin{tabular}{p{0.20\textwidth}p{0.74\textwidth}}
\toprule
\textbf{Prompt} & \textbf{Template and requirements} \\
\midrule

\textbf{Criticality Locator}
&
\textit{(intro)} Locate later plots and scenes that may contain evidence of the
long-term consequences of a candidate critical behavior.

\textit{(requirements)} Use only the supplied event ledger and later scenes. Do
not select the candidate's own scene, an earlier scene, or a scene that merely
repeats its wording or emotional tone. Return at most three supporting scene
identifiers and a concise explanation of the possible downstream consequence.
Return an empty list when no later evidence can be identified.
\\
\midrule

\textbf{Criticality Verifier}
&
\textit{(intro)} Independently verify whether a candidate behavior should retain
its critical-behavior label.

\textit{(requirements)} Keep a candidate only when the supplied later evidence
contains a concrete event, relationship change, decision, or consequence that
could plausibly differ under another reasonable choice. Every retained case must
cite at least one supporting scene and action identifier. Vague plot importance,
current-scene mood, or repeated wording does not constitute evidence. Return
\texttt{keep} or \texttt{downgrade} in strict JSON.
\\
\midrule

\textbf{Scene-Time Relabeling}
&
\textit{(intro)} Re-annotate the time at which the actions and dialogue in each
scene actually occur.

\textit{(requirements)} Use only explicit or directly grounded evidence from the
original plot text. A future appointment, a remembered event, or a date on a
document is not automatically the time of the current scene. If the scene time
is not established, return null for both the time label and its evidence.
\\
\midrule

\textbf{Timeline Anomaly Detection}
&
\textit{(intro)} Detect confident cases in which the story-time order of scenes
differs from their textual presentation order.

\textit{(requirements)} Treat textual order as the default timeline. Report only
scenes or contiguous scene blocks whose primary actions and dialogue clearly
occur earlier in story time. Do not move a scene merely because it contains a
recollection, discussion of a past event, or a future plan. Return proposed
moves and uncertain cases separately.
\\
\midrule

\textbf{Timeline Review}
&
\textit{(intro)} Independently review each proposed timeline adjustment against
the original plot text.

\textit{(requirements)} Check both the moved block and its insertion neighborhood.
Return \texttt{accept} only when the boundaries and insertion point are
supported, \texttt{reject} when the textual order should be preserved or
evidence is insufficient, and \texttt{modify} when a different move is required.
Provide source-grounded reasoning for every decision.
\\
\midrule

\textbf{Entity Resolution}
&
\textit{(intro)} Resolve source-level character names into persistent character
entities.

\textit{(requirements)} Group names only when the supplied time-ordered
appearances establish that they refer to the same person. Do not merge people
based only on shared surnames, titles, gender, or proximity. Every source name
must appear exactly once in an entity group or in the unresolved list. Prefer
conservative unresolved assignments to unsupported merges.
\\

\bottomrule
\end{tabular}%
}
\end{table*}

\begin{table*}[t]
\caption{Prompts for dynamic character-profile construction.}
\label{tab:prompt_profile}
\centering
\resizebox{\textwidth}{!}{%
\begin{tabular}{p{0.20\textwidth}p{0.74\textwidth}}
\toprule
\textbf{Prompt} & \textbf{Template and requirements} \\
\midrule

\textbf{Profile Initialization}
&
\textit{(intro)} Construct the initial dynamic profile of a character
immediately before the character's first recorded behavior.

\textit{(requirements)} Use only the pre-action character description. Do not
use the first action, later events, future relationships, or outside knowledge
of the novel. Summarize the character's supported identity, social position,
temperament, interaction style, current relationships, and relevant lived
experience in a compact profile. Return an evidence assessment together with
the profile.
\\
\midrule

\textbf{Profile Update}
&
\textit{(intro)} Update a character profile after a valid critical behavior and
the accumulated actions since the previous profile version.

\textit{(requirements)} Set \texttt{changed} to true only when the action window
causes a durable change in identity, social position, stable attitudes,
important relationships, goals, self-understanding, or relevant lived
experience. Repeated or temporary behavior does not require a new version. When
\texttt{changed=false}, return a null profile. When \texttt{changed=true},
rewrite the complete current profile rather than producing a chronological
action log. Do not use future events or counterfactual alternatives.
\\

\bottomrule
\end{tabular}%
}
\end{table*}

\begin{table*}[t]
\caption{Prompts for counterfactual rewriting and record-level anonymization.}
\label{tab:prompt_counterfactual}
\centering
\resizebox{\textwidth}{!}{%
\begin{tabular}{p{0.20\textwidth}p{0.74\textwidth}}
\toprule
\textbf{Prompt} & \textbf{Template and requirements} \\
\midrule

\textbf{Counterfactual Rewriting}
&
\textit{(intro)} Rewrite the initially generated counterfactual behaviors for a
source behavior.

\textit{(requirements)} The three alternatives must remain plausible responses
to the same scene while violating the target character's personality, values,
goals, or behavioral style. Match the real behavior in length, structural form,
speech format, register, and degree of formality. Preserve a quotation for
dialogue behaviors and avoid quotations for non-dialogue behaviors. The
alternatives must not be simple paraphrases and should differ along distinct
behavioral dimensions. Return exactly three independent strings in strict JSON.
\\
\midrule

\textbf{Record Anonymization}
&
\textit{(intro)} Replace person names in a dataset record with consistent
anonymous character identifiers.

\textit{(requirements)} Identify every person-name form appearing in the record,
including full names, nicknames, abbreviations, possessives, and
title--name combinations. Map forms referring to the same person to the same
\texttt{Character N}, ordered by first appearance. Preserve place names, animal
names, relational terms, structural identifiers, and labels required for
evaluation. Return a strict JSON name map.
\\

\bottomrule
\end{tabular}%
}
\end{table*}

\begin{table*}[t]
\caption{Shared character-context prompt used for RPO training.}
\label{tab:prompt_training}
\centering
\resizebox{\textwidth}{!}{%
\begin{tabular}{p{0.20\textwidth}p{0.74\textwidth}}
\toprule
\textbf{Prompt} & \textbf{Template and requirements} \\
\midrule

\textbf{Character Context}
&
\textit{(intro)} You are role-playing as the target character. Decide what the
character will do next based on the supplied character profile, memories, and
current scene.

\textit{(requirements)} Treat the provided Character Profile, Long-term Memory,
Recent Memory, and Current Scene as the complete and authoritative history for
the task. Do not rely on prior knowledge of the original novel, reconstruct
omitted events, or infer future developments. The profile describes the current
character state; long-term memory contains validated key behaviors; recent
memory provides local behavioral context; and the current scene contains the
situation immediately before the target behavior.

During training, the source behavior is provided as \texttt{chosen} and one
counterfactual behavior is provided as \texttt{rejected}. The model is trained
to prefer the source behavior while also increasing its conditional likelihood
under the auxiliary generation loss.
\\

\bottomrule
\end{tabular}%
}
\end{table*}

\begin{table*}[t]
\caption{Prompt used for four-way next-behavior selection and context ablations.}
\label{tab:prompt_evaluation}
\centering
\resizebox{\textwidth}{!}{%
\begin{tabular}{p{0.20\textwidth}p{0.74\textwidth}}
\toprule
\textbf{Prompt} & \textbf{Template and requirements} \\
\midrule

\textbf{Next-Behavior Selection}
&
\textit{(intro)} Choose the action that the target character would most likely
take next.

\textit{(requirements)} Use only the supplied Character Profile, Long-term
Memory, Recent Memory, and Current Scene. Do not use prior knowledge of the
original novel or any event that occurs after the target behavior. Four
candidate behaviors are provided, including one source behavior and three
counterfactual alternatives. Output exactly one letter: \texttt{A},
\texttt{B}, \texttt{C}, or \texttt{D}. Do not provide an explanation or any
additional text.
\\
\midrule

\textbf{Context Ablations}
&
\textit{(intro)} Evaluate the effect of individual character-information sources
on next-behavior selection.

\textit{(requirements)} Use the same selection prompt while selectively removing
the book identity, character profile, long-term memory, recent memory, or both
profile and memory. All other scene and option information remains unchanged.
The model must follow the same single-letter output constraint in every
condition.
\\

\bottomrule
\end{tabular}%
}
\end{table*}

\end{document}